\documentclass[conference]{IEEEtran}
\IEEEoverridecommandlockouts

\usepackage{booktabs}
\usepackage{multirow}
\usepackage{makecell}

\usepackage{cite}
\usepackage{amsmath,amssymb,amsfonts}
\usepackage{algorithmic}
\usepackage{graphicx}
\usepackage{textcomp}
\usepackage{xcolor}
\def\BibTeX{{\rm B\kern-.05em{\sc i\kern-.025em b}\kern-.08em
    T\kern-.1667em\lower.7ex\hbox{E}\kern-.125emX}}
\begin{document}

\title{KVPruner: Structural Pruning for Faster and Memory-Efficient Large Language Models\\}


\author{\IEEEauthorblockN{Bo Lv$^{1}$, Quan Zhou$^{1*}$, Xuanang Ding$^{1}$, Yan Wang$^{1}$, Zeming Ma$^{1}$}
\IEEEauthorblockA{
\textit{$^1$Huazhong University of Science and Technology} \\
\{bolv, quanzhou, dingxuanang, milliewang, mzm\}@hust.edu.cn}
\thanks{$^*$Corresponding author.}
}

\maketitle

\begin{abstract}
The bottleneck associated with the key-value(KV) cache presents a significant challenge during the inference processes of large language models.
While depth pruning accelerates inference, it requires extensive recovery training, which can take up to two weeks.
On the other hand, width pruning retains much of the performance but offers slight speed gains.
To tackle these challenges, we propose KVPruner to improve model efficiency while maintaining performance.
Our method uses global perplexity-based analysis to determine the importance ratio for each block and provides multiple strategies to prune non-essential KV channels within blocks.
Compared to the original model, KVPruner reduces runtime memory usage by 50\% and boosts throughput by over 35\%.
Additionally, our method requires only two hours of LoRA fine-tuning on small datasets to recover most of the performance.
\end{abstract}


\section{Introduction}
Large language models (LLMs) refer to natural language processing (NLP) models with a massive number of parameters \cite{10096429,10447215,10446156, DBLP:journals/corr/abs-1810-04805, vaswani2017attention}, commonly based on the Transformer architecture \cite{vaswani2017attention}.
These models have also found widespread applications in fields such as speech processing \cite{huang2018music, ren2019fastspeech, gong2022ssast} and computer vision \cite{dosovitskiy2020image, liu2021swin, caron2021emerging}.
In recent years, LLMs have demonstrated remarkable capabilities in handling complex tasks in applications like dialogue systems \cite{zheng2024judging, das2017visual} and knowledge-based question answering \cite{zhao2024expel, fan2024survey, lewis2020retrieval}, significantly accelerating the development of downstream applications.
However, as model sizes continue to grow, the challenges related to inference efficiency have become more pronounced.

Currently, optimization methods for large models include pruning (structured pruning \cite{ma2023llm, kim2024shortened, xia2023sheared} and unstructured pruning \cite{sun2023wanda,frantar2023sparsegpt}), quantization \cite{frantar2022gptq,lin2024awq,zhang2024pqcache}, and distillation \cite{wang2024rdrec, saad2023udapdr}.
This work focuses on structured pruning, making it more deployment-friendly and hardware-friendly.
In modern structured pruning algorithms for LLMs, LLM-Pruner \cite{ma2023llm} achieves model size reduction by removing inter-group dependencies in the network.
Sheared-LLaMA \cite{xia2023sheared} not only removes structures within groups but also prunes less important blocks to achieve compression.

\begin{figure}[htbp]
\centering
\includegraphics[width=0.48\textwidth,height=0.5\textheight,keepaspectratio]{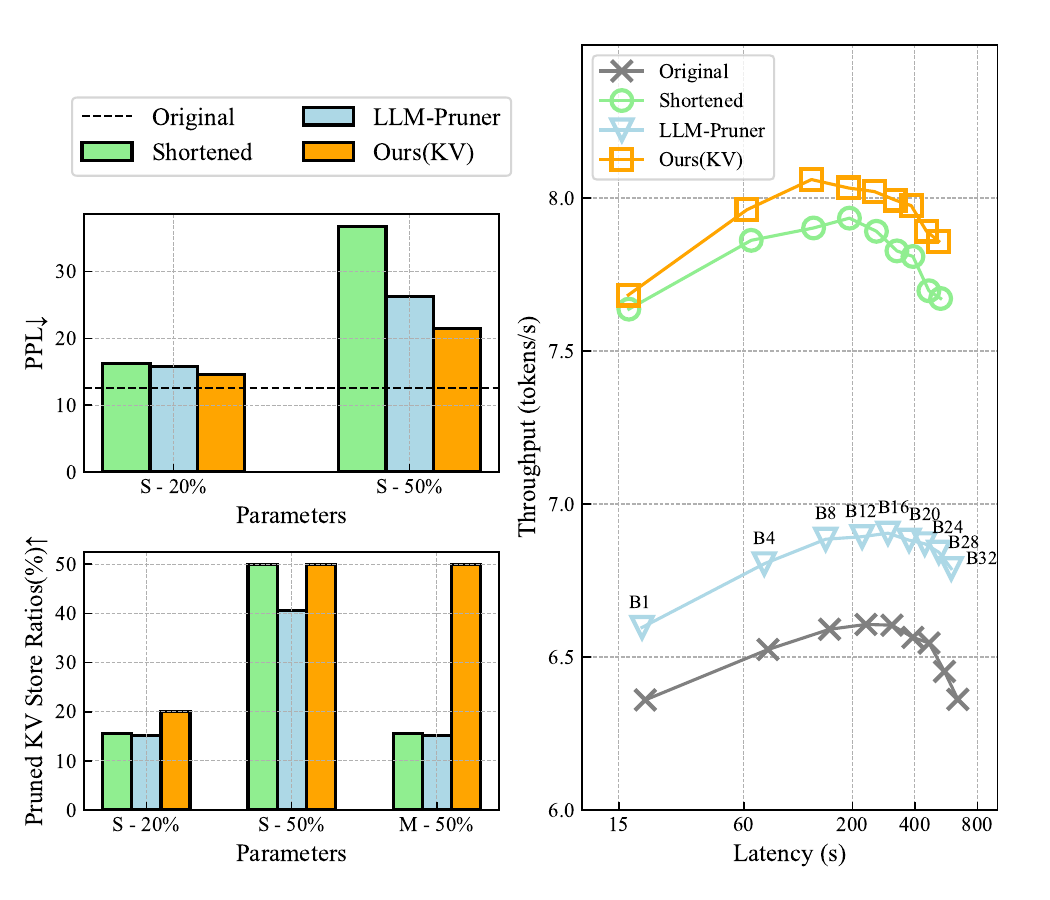}
\caption{The inference results of the pruned LLaMA-7B model on an NVIDIA A100 GPU, measuring computation latency and throughput with caching disabled.
Left side: The top left compares perplexity(PPL) across different strategies under the same pruning ratio and fine-tuning steps, where our method demonstrates superior performance.
The bottom left shows the key-value (KV) cache usage, where our approach achieves more significant KV memory pruning at both strategy-level and model parameter-level pruning ratios.
Right side: Under the same model parameter settings, KVPruner achieves faster inference speeds compared to Shortened-LLM \cite{kim2024shortened} and LLM-Pruner \cite{ma2023llm} pruning method.}
\label{fig1}
\end{figure}

These methods can employ LoRA \cite{hu2021lora} to recover performance, but the gains in runtime memory efficiency and inference performance are still minimal.
Shortened-LLM \cite{kim2024shortened} aggressively removes entire blocks to speed up inference, but it shows that directly removing blocks requires retraining with CPT \cite{kim2024shortened}, which can take several weeks to restore the pruned model.

Inference in autoregressive LLMs demands high computational resources and requires a KV cache to store previous tokens for generating new ones, increasing huge overhead on both memory and computation.
A typical scenario is in Retrieval-Augmented Generation (RAG) \cite{zhao2024expel, fan2024survey, lewis2020retrieval}, where large models must handle long prompts and generate extended responses.
Recent inference optimization models, such as SGLang \cite{zheng2023efficiently}, point out that current LLM inference bottlenecks arise from the need to cache and recompute KV entries.
On the GPU side, methods like FlashAttention \cite{dao2022flashattention, dao2023flashattention} reduce KV memory accesses during inference to increase throughput.
In large-scale inference scenarios, DistServe \cite{zhong2024distserve} also highlights that KV caching significantly impacts LLM inference throughput.

Therefore, in the context of structured pruning for large models, KV cache in transfomer is a crucial metric for evaluating runtime efficiency and throughput.
This study introduces KVPruner, which focuses on KV pruning for large models (e.g., LLaMA \cite{touvron2023llama}).
Under similar fine-tuning conditions, the pruned model achieves performance comparable to or better than previous structured pruning methods (as shown in Fig.~\ref{fig1}).
The main contributions of this work are as follows:

\begin{itemize}
    \item A structured pruning method for KV cache, which significantly reduces runtime memory usage.
    \item A perplexity(PPL)-based sensitivity analysis method for evaluating KV pruning at the block level, enabling optimal pruning ratios for each block.
    \item A method for intra-block evaluation of query (Q), key (K), value (V), and output (O) channels, validating the effectiveness of various pruning strategies.
\end{itemize}

\section{Methods}

The Decoder-Only architecture for large models, featuring the Transformer \cite{vaswani2017attention} decoder structure with multiple stacked decoders, exhibits exceptional performance.
Each decoder module primarily includes a normalization layer (Norm), multi-head attention (MHA), and a feedforward neural network (FFN).
The main computational and storage bottleneck lies in the MHA module.
Therefore, we focus on pruning KV cache to achieve better performance.
As shown in Fig.~\ref{fig2}, Our KVPruner method consists of two key phases: Phase A evaluates the optimal global pruning ratio for a given pruning target,
while Phase B handles the intra-block importance of Q, K, V and O channels and executes the pruning operation.
Our method performs one-shot pruning on the model, resulting in no additional overhead during runtime.

\subsection{ Global Pruning Ratio Awareness}\label{AA}

As shown in Fig.~\ref{fig3}, we analyze the dimensions of each block to ensure performance in both task-specific and generalization tasks.
We observe that the KV channels within different blocks have varying levels of impact on PPL in LLaMA.
Additionally, though the range of Q, K, and V parameters varies significantly from block to block, the overall trend aligns with the block-level PPL changes, with the first and last blocks showing the most significant impact.

\begin{figure}[htbp]
\centering
\includegraphics[width=0.4\textwidth,height=0.5\textheight,keepaspectratio]{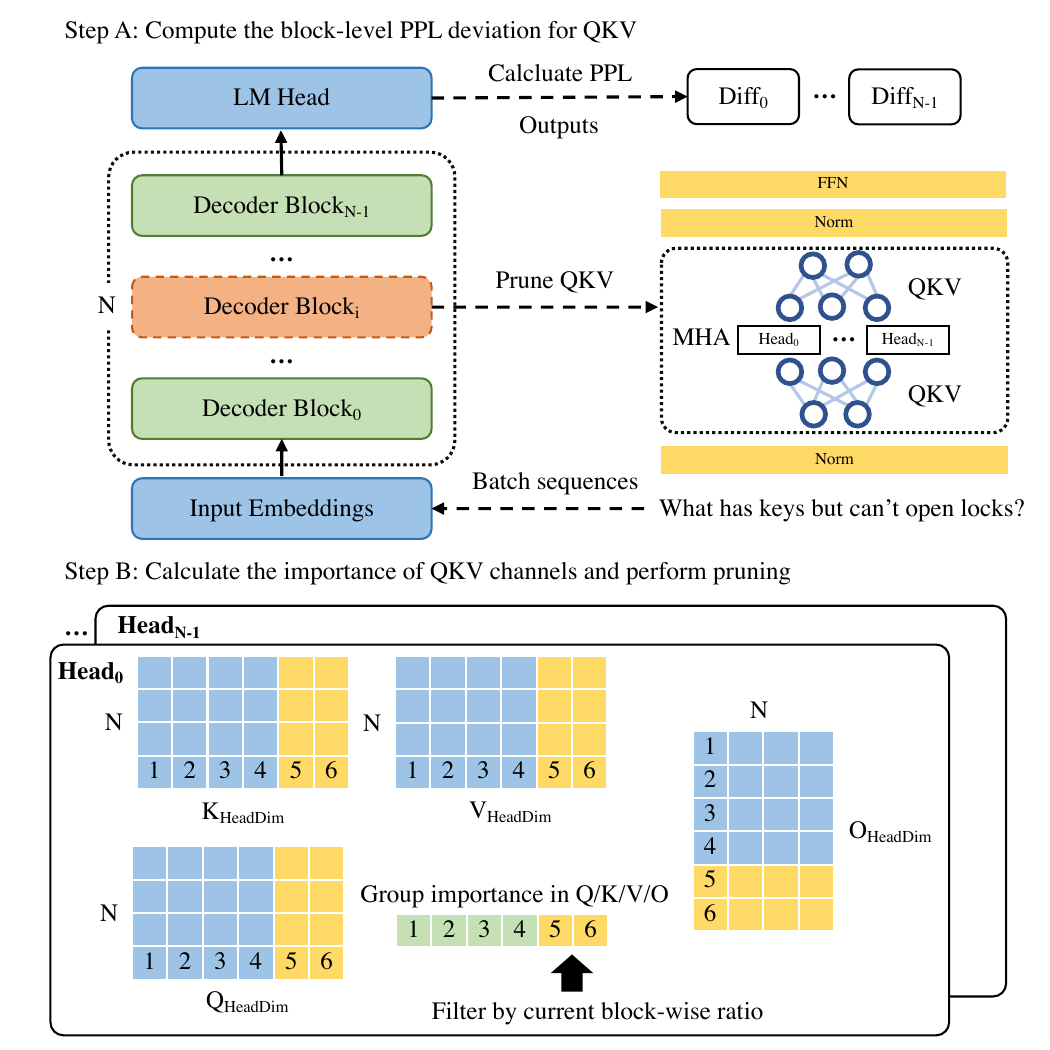}
\caption{Illustrates the simplified workflow of the KVPruner in LLMs.
The pruning process consists of two main steps:
First, global sensitivity analysis assigns the optimal pruning ratio to each block.
Second, local channel sensitivity aggregates the importance of Q, K, V and O channels for evaluation and removes the less important ones.
After completing these steps, LoRA is applied to quickly recovery the performance.}
\label{fig2}
\end{figure}

Therefore, we define the PPL change for each block as $\Delta \text{PPL}_i$, which allows us to compute the global importance ratio or priority for each block.
As shown in step A of Fig.~\ref{fig2}, we propose two methods for calculating the block pruning ratio distribution and further compute the sensitivity distribution within each block.

\subsubsection{PPL-based Sensitivity Distribution}

\begin{figure}[htbp]
\centering
\includegraphics[width=0.45\textwidth,height=0.45\textheight,keepaspectratio]{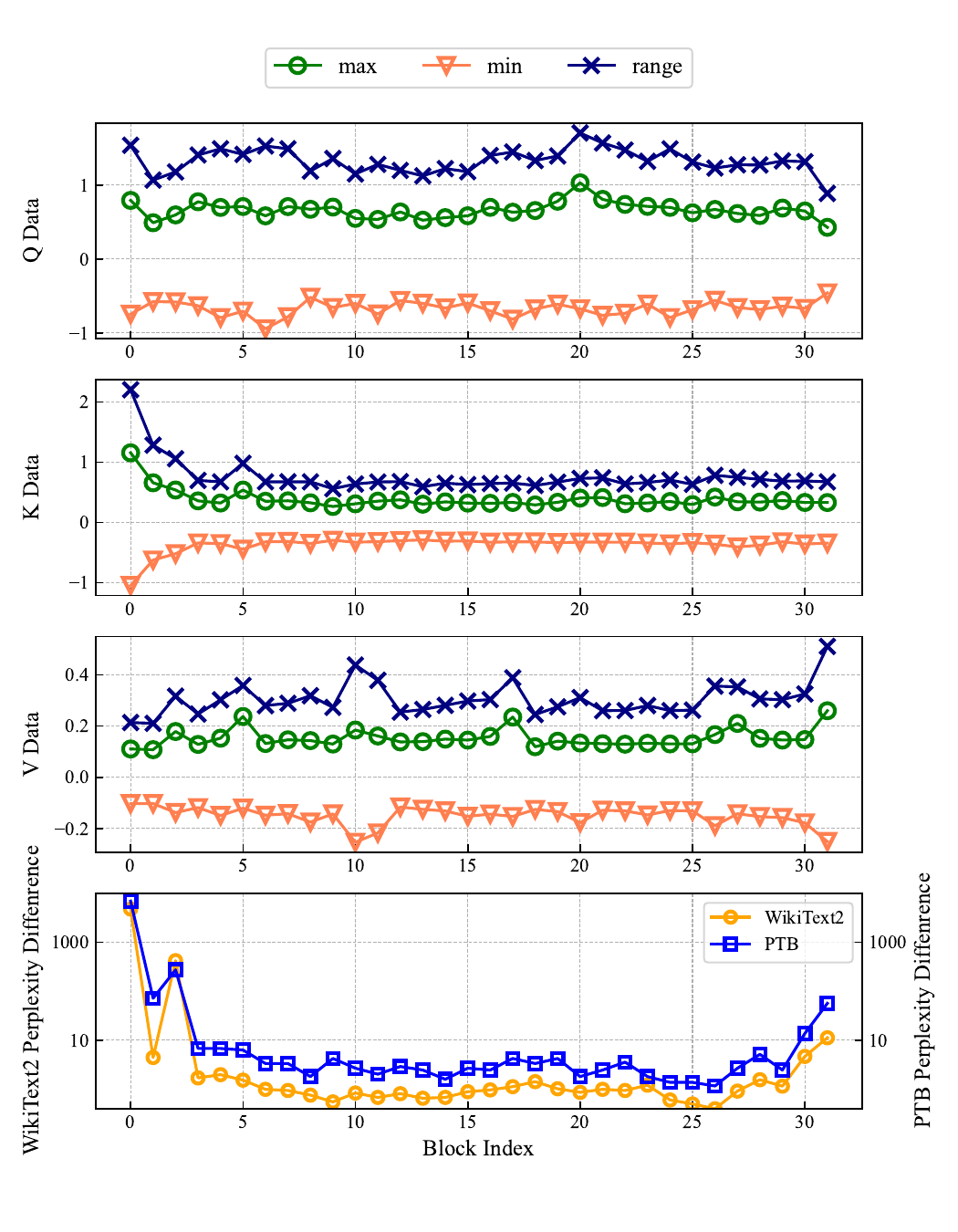}
\caption{The first three graphs show the range of QKV weight changes in a specific block.
Estimated on the evaluation set, the fourth graph illustrates the sensitivity changes after removing the KV from the block.}
\label{fig3}
\end{figure}

This method allocates the pruning ratio for each block based on its relative PPL changes, as shown in Fig.~\ref{fig3}.
The key idea is to assign lower pruning ratios to blocks with higher sensitivity, as indicated by larger PPL changes. Let \(\Delta \text{PPL}_i\) represent the PPL changes for the \(i\)-th block. The pruning ratio for each block is computed by taking the inverse of the exponential change in PPL, normalized by the total exponential change across all blocks. The pruning ratio for the \(i\)-th block is given by Eq.~\eqref{e1}.

\begin{equation}
P_i = \frac{\frac{1}{e^{\Delta \text{PPL}_i} + \epsilon}}{\sum_{j=1}^{N} \frac{1}{e^{\Delta \text{PPL}_j} + \epsilon}} \cdot P_{\text{total}}\label{e1}
\end{equation}

where \(P_i\) is the pruning ratio for the \(i\)-th block, \(\Delta \text{PPL}_i\) is the PPL changes, \(e^{\Delta \text{PPL}_i}\) represents the exponential change, \(\epsilon\) is a small constant to avoid division by zero, \(P_{\text{total}}\) is the global pruning ratio, and \(N\) is the total number of blocks.

\subsubsection{Rank-based Sensitivity Distribution}

In this method, we prioritize blocks based on their PPL changes.
Blocks are sorted by their PPL change, and pruning ratios are assigned according to this sorted order.
Blocks with higher priorities are pruned, while those with lower priorities are not.

Let \(id_i\) be the ranking of the \(i\)-th block based on its PPL change.
We define a threshold based on the total number of blocks and the total pruning ratio.
The pruning ratio for each block is then calculated as shown in Eq.~\eqref{e2}.

\begin{equation}
P_i =
\begin{cases}
1, & \text{if } id_i \leq \lceil P_{\text{total}} \cdot N \rceil \\
0, & \text{otherwise}
\label{e2}
\end{cases}
\end{equation}

where \(P_i\) is the pruning ratio for the \(i\)-th block, \(id_i\) is the rank of the \(i\)-th block according to its PPL change, \(P_{\text{total}}\) is the total global pruning ratio, and \(N\) is the total number of blocks.

\subsection{Local Channel Importance Evaluation}

After determining the block-level KV pruning ratio in Section~\ref{AA},
we further evaluate the intra-block importance of Q, K, V and O matrix channels, as shown in step B of Fig.~\ref{fig2}.
First, we need to compute the total score for all elements in each channel, then we provide three methods to calculate the importance score for each channel.
These methods are used to evaluate the significance of each channel, which can later be utilized for pruning purposes.
Let \( W \) be the weight matrix corresponding to Q, K, V and O channels with dimensions \( (C_{\text{in}}, C_{\text{out}}) \), where \( C_{\text{in}} \) and \( C_{\text{out}} \) are the number of input and output channels, respectively.

\subsubsection{L1 Method}

L1 is a basic metric commonly used in pruning literature.
It measures the importance of each channel by computing the sum of the absolute values of the weights within that channel.
For the \( i \)-th channel, the L1 importance score \( S_{\text{L1},i} \) is computed as shown in Eq.~\eqref{e3}.

\begin{equation}
S_{\text{L1},i} = \sum_{j=1}^{C_{\text{in}}} \sum_{k=1}^{C_{\text{out}}} \left| W_{i,j,k} \right|\label{e3}
\end{equation}

where the weight element \( W_{i,j,k} \) represents the weight for the \( i \)-th channel. \( C_{\text{in}} \) and \( C_{\text{out}} \) denote the input and output channel dimensions, respectively.

\subsubsection{L2 Method}

The L2 method extends the L1 metric by squaring the weights, which emphasizes larger weights.
The L2 importance score \( S_{\text{L2},i} \) for the \( i \)-th channel is computed as shown in Eq.~\eqref{e4}.

\begin{equation}
S_{\text{L2},i} = \sum_{j=1}^{C_{\text{in}}} \sum_{k=1}^{C_{\text{out}}} \left( W_{i,j,k} \right)^2 \label{e4}
\end{equation}

\subsubsection{Taylor Method}

The Taylor method leverages the error on a calibration dataset and uses the first-order term of the Taylor expansion to approximate the importance of each channel.
For a given calibration dataset \( D \), we compute the gradient of the loss function \( \mathcal{L} \) with respect to the weights.
The Taylor importance score \( S_{\text{Taylor},i} \) for the \( i \)-th channel is computed as shown in Eq.~\eqref{e5}.

\begin{equation}
S_{\text{Taylor},i} = \sum_{j=1}^{C_{\text{in}}} \sum_{k=1}^{C_{\text{out}}} \left| \frac{\partial \mathcal{L}}{\partial W_{i,j,k}} W_{i,j,k} \right| \label{e5}
\end{equation}

where \( \mathcal{L} \) is the training loss, and \( \frac{\partial \mathcal{L}}{\partial W_{i,j,k}} \) is the gradient of the loss with respect to the weight \( W_{i,j,k} \).

The importance score of a channel in the KV dimension is the average of the scores for the corresponding Q, K, V and O channels, calculated using the following formula:

Let \( M \) represent the method used to calculate the importance score (either L1, L2, or Taylor). For the \(i\)-th channel, the average importance score across the Q, K, V, and O matrices is given by the following unified formula as shown in Eq.~\eqref{e6}.

\begin{equation}
S_{M,i}^{\text{avg}} = \frac{1}{4} \left( S_{M,i}^{Q} + S_{M,i}^{K} + S_{M,i}^{V} + S_{M,i}^{O} \right) \label{e6}
\end{equation}

where \( S_{M,i}^{Q} \) is the importance score of the \(i\)-th channel in the Q matrix using method \( M \), \( S_{M,i}^{K} \) is the importance score of the \(i\)-th channel in the K matrix using method \( M \), \( S_{M,i}^{V} \) is the importance score of the \(i\)-th channel in the V matrix using method \( M \), and \( S_{M,i}^{O} \) is the importance score of the \(i\)-th channel in the O matrix using method \( M \).

\subsection{Fine-Tuning After Pruning}

For small-scale structured pruning, our method causes minimal change in PPL, but fine-tuning can still be applied to further improve model performance at a low cost.
Compared to the Shortened-LLM \cite{kim2024shortened} method which requires high-cost CPT \cite{kim2024shortened} retraining on H100 GPUs for two weeks, our approach only requires a few hours of fine-tuning using LoRA \cite{hu2021lora}, significantly reducing training costs.

\section{Experiments}

\subsection{Experimental Setup}

To evaluate the performance of our method compared to existing structured pruning methods in terms of memory usage, inference latency, throughput, and PPL on the test set, we selected the LLaMA-7B \cite{touvron2023llama} model as our base model.
Our baseline methods include LLM-Pruner \cite{ma2023llm} and Shortened-LLM \cite{kim2024shortened}, representing state-of-the-art width and depth structured pruning approaches, respectively.

\textbf{PPL Evaluation:} We assess block-level KV sensitivity by randomly selecting a portion of the WikiText2 dataset for screening and evaluating PPL performance on the remaining WikiText2 and PTB datasets.

\textbf{Inference Overhead Evaluation:} To avoid the storage bottleneck mentioned in Shortened-LLM \cite{kim2024shortened}, we disable inference caching during the evaluation, calculating inference cost purely based on computation. This evaluation applies to scenarios where computation and storage are decoupled, such as in a separated architecture \cite{hu2024memserve,zhong2024distserve, qin2024mooncake}.

\textbf{Latency and Throughput Evaluation:} Following the evaluation standard of Shortened-LLM \cite{kim2024shortened}, we preheat the model 10 times and then measure the time \( T \) for a given batch size and output sequence length \( L \). We calculate throughput as \( ML/T \), where the input token size is 1024.

\textbf{Hardware Environment:} We perform pruning and two-batches LoRA fine-tuning on an NVIDIA A100 GPU, and test inference latency and throughput on both NVIDIA A100 GPU and NVIDIA RTX 3090.

\begin{table}[]
\caption{PPL values under different pruning methods and pruning ratios.}
\centering
\renewcommand{\arraystretch}{0.9}
\setlength\tabcolsep{4pt}
\begin{tabular}{cccccc}
	\toprule
	\multirow{2}{*}{\textbf{Method}} &
	\textbf{\makecell{Ratio}} &
	\textbf{\makecell{Parameters}} &
	\textbf{\makecell{kv store↓}} &
	\multicolumn{2}{c}{\textbf{\makecell{PPL↓(LoRA)}}}\\&
	\textbf{\makecell{(\%)}} &
	\textbf{\makecell{(B)}} &
	\textbf{\makecell{(GB)}} &
	\textbf{WikiText2} &
	\textbf{PTB} \\
	\midrule
	LLaMA-7B & 0 & 7 & 8 & 12.61 & 22.14 \\
	LLM-Pruner & 20 & 5.5 & 6.78 & 15.73 & 71.48 \\
	LLM-Pruner & 50 & 3.5 & 4.75 & 26.24 & 112.02 \\
	Shortened & 20 & 5.5 & 6.75 & 16.29 & 70.37 \\
	Shortened & 50 & 3.5 & \textbf{4} & 36.79 & 124.97 \\
	Ours & 20(qkv) & 6.3 & 6.4 & \textbf{14.55} & \textbf{27.18} \\
	Ours & 50(qkv) & 5.5 & \textbf{4} & 21.5 & 40.33 \\
	\bottomrule
\end{tabular}
\label{ppl table}
\end{table}

\begin{table}[]
\caption{Latency and throughput values under different pruning methods and pruned model parameters.}
\centering
\renewcommand{\arraystretch}{0.9}
\setlength\tabcolsep{1pt}
\begin{tabular}{cccccc}
	\toprule
	\multirow{3}{*}{\textbf{Method}} &
	\multirow{3}{*}{\textbf{\makecell{Parameters\\(B)}}} &
	\multicolumn{2}{c}{\textbf{\makecell{A100 40GB}}}&
	\multicolumn{2}{c}{\textbf{\makecell{RTX 3090 24GB}}}\\&
	&
	\textbf{Latency↓} &
	\textbf{Throughput↑} &
	\textbf{Latency↓} &
	\textbf{Throughput↑} \\&
	&
	\textbf{(s)} &
	\textbf{(tokens/s)} &
	\textbf{(s)} &
	\textbf{(tokens/s)}  \\
	\midrule
	LLaMA-7B & 7 & 64.995 & 63.02 & 75.804 & 27.017 \\
	LLM-Pruner & 5.5 & 64.045 & 63.955 & 66.187 & 30.943 \\
	Shortened & 5.5 & 54.425 & 75.26 & 63.759 & 32.122 \\
	Ours & 6.3 & 60.447 & 67.762 & 62.729 & 32.649 \\
	Ours & 5.5 & \textbf{52.438} & \textbf{78.111} & \textbf{58.132} & \textbf{35.23 }\\
	\bottomrule
\end{tabular}
\label{latency table}
\end{table}

\subsection{Main Results}

Table~\ref{ppl table} and Table~\ref{latency table} show the performance comparison under 20\% and 50\% pruning ratios.
When the remaining number of model parameters is the same after pruning, our method reduces runtime memory usage by 50\%, significantly surpassing other methods.
Additionally, our approach demonstrated superior inference speed and throughput on both A100 and 3090 machines.
Under the same 20\% and 50\% pruning ratios, our method also achieved lower PPL than other methods, indicating better performance.
We also observe that while Shortened-LLM approaches the performance of our method under higher pruning ratios, its coarse pruning approach prevented effective recovery within a reasonable fine-tuning time.

\begin{table}[]
\caption{Comparison of pruning strategies with a KV pruning ratio of 20\%.}
\centering
\renewcommand{\arraystretch}{0.9}
\setlength\tabcolsep{1pt}
\begin{tabular}{cccccc}
	\toprule
	\multirow{2}{*}{\textbf{Global}} &
	\multirow{2}{*}{\textbf{\makecell{Block}}} &
	\multicolumn{2}{c}{\textbf{\makecell{PPL↓(Pruned)}}}&
	\multicolumn{2}{c}{\textbf{\makecell{PPL↓(LoRA)}}}\\&
	&
	\textbf{WikiText2} &
	\textbf{PTB} &
	\textbf{WikiText2} &
	\textbf{PTB} \\
	\midrule
	\multirow{4}{*}{Uniform} & 01 & 17.89 & 32.81 & 15.46 & 64.83 \\
	& l1 & 768.56 & 1010.26 & 245.64 & 327.98 \\
	& l2 & 860.75 & 891.55 & 22.75 & 43.02 \\
	& taylor & 564.49 & 566.70 & 24.41 & 45.88 \\
	\midrule
   	\multirow{4}{*}{PPL-based} & 01 & \textbf{15.04} & \textbf{29.33} & \textbf{14.55} & \textbf{27.18} \\
	& l1 & 178.32 & 230.76 & 20.68 & 36.65 \\
	& l2 & 421.14 & 434.50 & 20.48 & 36.58 \\
	& taylor & 129.44 & 143.84 & 18.18 & 33.18 \\
	\bottomrule
\end{tabular}
\label{ablation_20 table}
\end{table}

\begin{table}[]
\caption{Comparison of pruning strategies with a KV pruning ratio of 50\%.}
\centering
\renewcommand{\arraystretch}{0.9}
\setlength\tabcolsep{1pt}
\begin{tabular}{cccccc}
	\toprule
	\multirow{2}{*}{\textbf{Global}} &
	\multirow{2}{*}{\textbf{\makecell{Block}}} &
	\multicolumn{2}{c}{\textbf{\makecell{PPL↓(Pruned)}}}&
	\multicolumn{2}{c}{\textbf{\makecell{PPL↓(LoRA)}}}\\&
	&
	\textbf{WikiText2} &
	\textbf{PTB} &
	\textbf{WikiText2} &
	\textbf{PTB} \\
	\midrule
	\multirow{4}{*}{Uniform} & 01 & 86.56 & 125.46 & 24.46 & 53.85 \\
	& l1 & 1939.74 & 2539.80 & 57.89 & 121.13 \\
	& l2 & 1552.55 & 1691.87 & 55.89 & 113.79 \\
	& taylor & 1272.11 & 1564.73 & 88.62 & 186.15 \\
	\midrule
   	\multirow{4}{*}{PPL-based} & 01 & \textbf{52.40} & \textbf{115.46} & \textbf{21.50} & \textbf{40.33} \\
	& l1 & 141.06 & 194.32 & 23.89 & 44.82 \\
	& l2 & 124.48 & 192.06 & 22.80 & 43.61 \\
	& taylor & 157.36 & 201.27 & 23.25 & 42.93 \\
	\bottomrule
\end{tabular}
\label{ablation_50 table}
\end{table}

\subsection{Ablation Study}

We perform a detailed analysis of several metrics mentioned in the method.
Table~\ref{ablation_20 table} and Table~\ref{ablation_50 table} present comparisons of global metrics and intra-block channel importance under 20\% and 50\% pruning ratios, respectively.
For the global metrics, we compare uniform pruning (i.e., without distinguishing blocks) with PPL sensitivity-based pruning and find that the latter achieves better results.
In the uniform pruning method (01, which selects based on ranked proportions), sensitive blocks with significant PPL changes are skipped, leading to better performance.
However, other pruning methods that prune sensitive blocks experience a sharp increase in PPL.
We also compare the 01, L1, L2, and Taylor pruning methods.
The 01 method performs well at small pruning ratios by avoiding PPL-sensitive blocks,achieving strong performance even before fine-tuning.
In other scenarios, the Taylor method performs best, followed by L2, with L1 showing the weakest performance.
However, all methods are able to recover to good levels under the same fine-tuning conditions, offering more options for pruning in specific task scenarios.

\section{Conclusion}

In this work, we propose KVPruner, a structured pruning method optimized for runtime KV cache during LLMs inference.
By combining global sensitivity awareness with local channel sensitivity techniques, KVPruner improves model efficiency while maintaining performance.
Unlike deep structured pruning methods, KVPruner recovers high performance in only two hours of LoRA fine-tuning on small datasets.
Our experiments on the LLaMA-7B model demonstrate that KVPruner reduces KV memory usage by 50\% and boosts throughput by over 35\%, offering a balanced and effective approach to pruning non-essential KV channels,
demostrating its ability to significantly reduce KV memory usage and improve inference efficiency.

\bibliographystyle{IEEEtran}
\bibliography{references}

\vspace{12pt}

\end{document}